\documentclass[10pt,journal,compsoc]{IEEEtran}
\ifCLASSOPTIONcompsoc
  \usepackage[nocompress]{cite}
\else
  \usepackage{cite}
\fi
\ifCLASSINFOpdf
  \usepackage[pdftex]{graphicx}
  \DeclareGraphicsExtensions{.pdf}
\else
\fi
\usepackage{amsmath}
\usepackage{amssymb}
\usepackage{algorithm}
\usepackage{algorithmic}

\usepackage{multirow}
\usepackage{booktabs}
\usepackage{amsfonts}
\usepackage{amsthm}

\begin{document}
%
\title{iCapsNets: Towards Interpretable Capsule Networks for Text Classification}
%
%
%
%

\author{Zhengyang~Wang,
    Xia~Hu,
    and~Shuiwang~Ji,~\IEEEmembership{Senior~Member,~IEEE}
    \IEEEcompsocitemizethanks{\IEEEcompsocthanksitem Zhengyang Wang, Xia Hu, and Shuiwang Ji are with the Department of Computer Science and Engineering, Texas A\&M University, College Station, TX, 77843.\protect\\
E-mail: sji@tamu.edu}
\thanks{Manuscript received May, 2020.}}

%
%

\markboth{IEEE Transactions on Pattern Analysis and Machine Intelligence,~Vol.~xx, No.~x, May~2020}%
{Wang \MakeLowercase{\textit{et al.}}: iCapsNets: Towards Interpretable Capsule Networks for Text Classification}
%



\IEEEtitleabstractindextext{%
\begin{abstract}
Many text classification applications require models with satisfying performance as well as good interpretability. Traditional machine learning methods are easy to interpret but have low accuracies. The development of deep learning models boosts the performance significantly. However, deep learning models are typically hard to interpret. In this work, we propose interpretable capsule networks~(iCapsNets) to bridge this gap. iCapsNets use capsules to model semantic meanings and explore novel methods to increase interpretability. The design of iCapsNets is consistent with human intuition and enables it to produce human-understandable interpretation results. Notably, iCapsNets can be interpreted both locally and globally. In terms of local interpretability, iCapsNets offer a simple yet effective method to explain the predictions for each data sample. On the other hand, iCapsNets explore a novel way to explain the model's general behavior, achieving global interpretability. Experimental studies show that our iCapsNets yield meaningful local and global interpretation results, without suffering from significant performance loss compared to non-interpretable methods.
\end{abstract}

}

\maketitle

\IEEEdisplaynontitleabstractindextext

%
\IEEEpeerreviewmaketitle

\IEEEraisesectionheading{\section{Introduction}\label{sec:intro}}

Text classification is an important task in natural language
processing~(NLP) research. With different predefined categorical
labels, models for text classification have various applications,
including sentiment analysis, topic categorization, and ontology
extraction~\cite{zhang2015character}.
A considerable body of efforts have been devoted to
developing machine learning models for text classification and many
successful models have been studied. However, as many practical
applications raise the requirement for interpretable models~\cite{du2018techniques,du2019learning,du2018towards,yang2019xfake,shu2019defend}, existing models
have not achieved a good trade-off between accuracy and interpretability.


Traditional text classifiers typically rely on statistical methods
like bag-of-words and bag-of-n-grams~\cite{joachims1998text}. By
simply counting the occurrences of words or n-grams and applying
machine learning methods like support vector machines, these methods
have achieved some success. However, without semantic understanding
of words or n-grams, the success is limited.


The development of distributed
representations~\cite{mikolov2013efficient,mikolov2013distributed,mikolov2013linguistic}
provides an effective way to model semantic meanings of words through word embeddings.
It has motivated applications of deep learning models on many NLP
tasks~\cite{hochreiter1997long,sutskever2014sequence,bahdanau2014neural,bengio2003neural,collobert2011natural,mikolov2010recurrent,chung2014empirical}.
Pre-trained word embeddings, like
word2vec~\cite{mikolov2013distributed} and
GloVe~\cite{pennington2014glove}, are made publicly available to
accelerate the research of deep learning on NLP.
In addition, other levels of text embeddings, such as
character embeddings~\cite{dos2014deep,zhang2015character,conneau2017very,xiao2016efficient},
sub-word embeddings~\cite{wu2016google},
and region embeddings~\cite{johnson2015effective,qiao2018a},
have also been explored.

Based on these embeddings, deep learning models based on recurrent neural
networks~(RNNs)~\cite{tang2015document,yogatama2017generative} and
convolutional neural
networks (CNNs)~\cite{johnson2015effective,johnson2015semi,johnson2017deep,kalchbrenner2014convolutional,kim2014convolutional}
have been extensively studied. While variants of RNNs, such as long
short-term memory~(LSTM)~\cite{hochreiter1997long} and gated
recurrent unit~(GRU)~\cite{chung2014empirical}, are known to be
effective for processing sequential data like text, many studies have
shown that CNNs are comparable with RNNs on NLP tasks.
The attention mechanism~\cite{bahdanau2014neural} is another important model.
On one hand, combining it with CNNs and RNNs
results in significant performance boost~\cite{yang2016hierarchical,wu2016google}.
On the other hand, the attention mechanism can be used as an alternative to CNNs and RNNs to
build deep learning models and set the record on various NLP tasks~\cite{vaswani2017attention,devlin2018bert}.
In terms of the accuracy, deep learning models usually outperform traditional text classifiers.
However, complex deep learning models work like black boxes and are hard to interpret~\cite{rai2020explainable}.


Distributed representations can also be combined with interpretable traditional machine learning method, resulting in simpler yet effective models. For example, FastText~\cite{joulin2017bag} combines
the traditional bag-of-words method with word embeddings, and uses the linear regression to perform classification.
It achieves comparable classification accuracies with complex deep learning models and is much more efficient in terms of memory usage
and computation speed. While the linear regression models can be interpreted, FastText applies an average/sum operation
to generate sentence embeddings from word embeddings, preventing it from telling which words are more important.

\begin{figure*}[t]
	\centering
	\includegraphics[width=\textwidth]{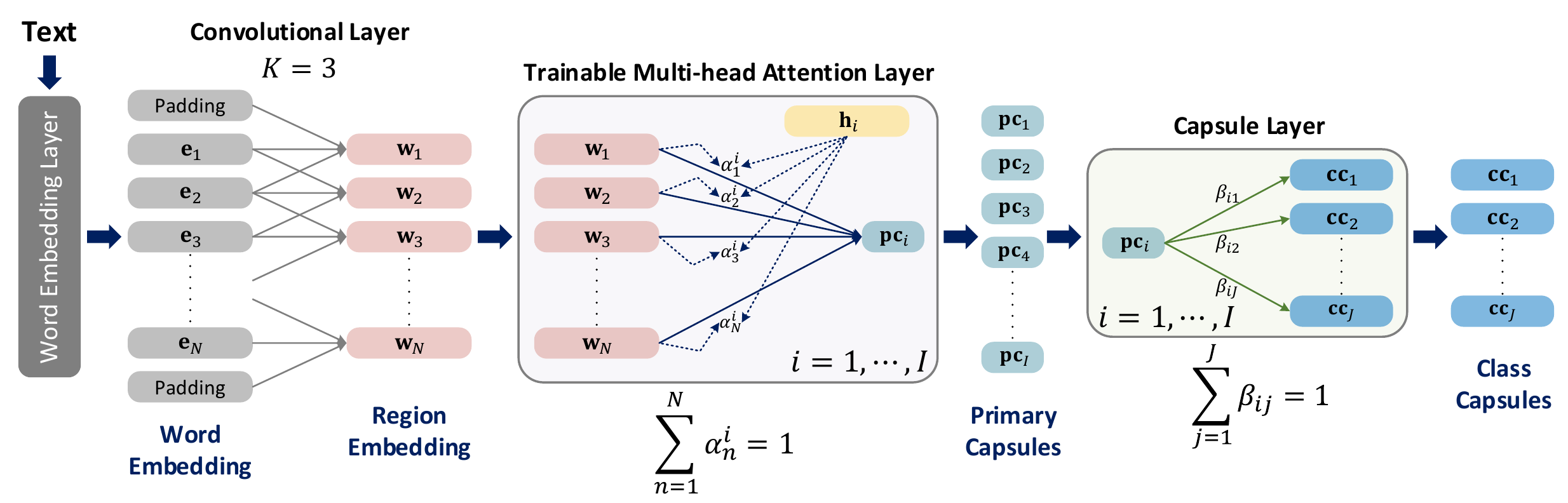}
	\caption{An illustration of the architecture of iCapsNets. From right to left are the capsule layer, the trainable multi-head attention layer, the 1D convolutional layer, and the word embedding layer. Note that the trainable multi-head attention layer can be replaced by the trainable multi-head hierarchical attention layer for long documents, as illustrated in Section~\ref{sec:hierarchical_attention}.}
	\label{fig:architecture}
\end{figure*}


In this work, we focus on developing interpretable deep learning models for text classification. Specifically, there are kinds of interpretabilities in terms of explaining machine learning models~\cite{du2018techniques,kopitar2019local}. 
The first one is the local interpretability, represents the ability of explaining why a specific decision is made with respect to a specific data sample. In contrast, the global interpretability, refers to the ability of showing how the model works generally, with respect to the whole dataset.
We aim at achieving both kinds of interpretabilities at the same time, without hurting the performance significantly.


In this work, we develop a novel deep learning model for text classification, which can be interpreted both locally and globally. Specifically, we extend the CapsNets~\cite{sabour2017dynamic} from computer vision tasks to text classification tasks, and make the following contributions:
\begin{itemize}
	\item We propose interpretable capsule networks~(iCapsNets) for text classification. To the best of our knowledge, this is the first work that achieves both local and global interpretability on CapsNets. 
	\item In iCapsNets, interpretation of classification results with respect to each data sample as well as the model's general behavior can be both obtained through novel, simple yet effective ways.
	\item Experimental results show that our iCapsNets yield meaningful interpretation results while having competitive accuracies compared to non-interpretable models, achieving a better trade-off between accuracy and interpretability.
\end{itemize}

\section{iCapsNets}\label{sec:model}

In this section, we first discuss the intuition behind the design of iCapsNets in Section~\ref{sec:motivation}. Then we introduce the overall architecture of iCapsNets in Section~\ref{sec:architecture}, followed by details of important layers in Sections~\ref{sec:attention}, \ref{sec:capsule} and~\ref{sec:hierarchical_attention}. Then we explain how to perform local and global interpretation of iCapsNets in Section~\ref{sec:interpreatation}.

\subsection{From CapsNets to iCapsNets}\label{sec:motivation}

CapsNets~\cite{sabour2017dynamic} employ capsules as the input and output of a layer and proposes the dynamic routing algorithm to perform the computation between capsules. A capsule is a vector, whose elements are the instantiation parameters of a specific type of entity. CapsNets were designed for computer vision tasks, where entities may refer to objects or object parts. In CapsNets, capsules are always normalized by a non-linear $Squash$ function so that the norm lies in $[0,1)$. The norm of a capsule, in turn, represents the probability that the corresponding entity is present in the input image. A capsule is said to be activated if the norm is close to 1. To understand how a capsule layer works, suppose we are given capsules that represent low-level entities; e.g., eyes, ears, and nose in the task of face detection. A capsule layer employs the dynamic routing algorithm to compute capsules that represent high-level entities, which can be faces in this case. Concretely, if low-level capsules indicate that two eyes and one nose are detected in the image, the dynamic routing algorithm is able to decide whether they belong to the same face and choose to activate the high-level capsule for face accordingly.

While convolutional neural networks~(CNNs) have become the dominant approach for classification tasks, CapsNets have two main advantages over CNNs. First, using vector-output capsules avoids the exponential inefficiencies incurred by replicating scalar-output feature detectors on a grid~\cite{hinton2011transforming}. Second, the dynamic routing algorithm is more effective than max-pooling, and allows neurons in one layer to ignore all but the most active feature detector in a local pool in the layer below~\cite{sabour2017dynamic}. Replacing stacked convolutional layers and max-pooling layers in CNNs with capsule layers results in the more efficient and effective CapsNets.

In this work, we extend CapsNets to text classification tasks and develop iCapsNets. As it is necessary for accurate text classification models to semantically understand texts, we propose to use capsules to represent different semantic meanings. Intuitively, there is a clear hierarchy between semantic meanings in text classification tasks. Taking topic categorization as an example, detecting certain basketball-related phrases will strongly suggest that a document be categorized into the ``sports'' topic. Here, ``basketball'' is a low-level semantic entity as opposed to ``sports''. With such intuitions, iCapsNets are designed to first capture low-level semantic entities~(primary capsules) from the entire input texts and then use the dynamic routing algorithm to compute high-level semantic entities~(class capsules), which correspond to predefined classes. Note that, given a sentence or document to classify, prior deep learning models usually generate a single sentence or document vector embedding which is fed into classifiers. In contrast, iCapsNets produce several primary capsules from the sentence or document, where each primary capsule focuses on capturing a specific semantic meaning.

\subsection{The Architecture of iCapsNets}\label{sec:architecture}

An illustration of the overall architecture of our iCapsNets is provided in Figure~\ref{fig:architecture}. As introduced above, the top layer of iCapsNets is a capsule layer, whose inputs are $I$ $d_p$-dimensional primary capsules ($\mathbf{pc}$) representing $I$ distinct low-level semantic meanings extracted from the whole input texts. The outputs are $J$ $d_c$-dimensional class capsules ($\mathbf{cc}$) representing high-level semantic meanings, where $J$ corresponds to the number of classes of the task. We apply the original dynamic routing algorithm~\cite{sabour2017dynamic}. In addition, we propose a weight sharing regularization to improve the efficiency and interpretability, which is discussed in detail in Section~\ref{sec:capsule}.

In the following discussions, we assume that the input to iCapsNets is a single sentence of $N$ words. For short documents, it is reasonable to concatenate the sentences into a single one. We discuss how to adapt iCapsNets for long documents in Section~\ref{sec:hierarchical_attention}.


To generate the primary capsules from input texts, iCapsNets start with a word embedding layer. While using character embeddings may lead to higher accuracies~\cite{zhang2015character}, it treats text as a kind of raw signal at character level, which is not helpful in achieving human-understandable interpretation results. The input sentence is transformed into a sequence of word embeddings $\mathbf{e}_1, \mathbf{e}_2, \cdots, \mathbf{e}_N \in \mathbb{R}^{d_e}$ through the word embedding layer. To be specific, given a vocabulary of size $V$ built according to the training dataset, the word embedding layer is essentially a $V \times h_e$ look-up table~\cite{collobert2011natural}. Here, $h_e$ is the dimension of word embedding space.


A 1-D convolutional layer with a kernel size of $K$ then transforms the word embeddings $\mathbf{e}_1, \mathbf{e}_2, \cdots, \mathbf{e}_N$ into region embeddings $\mathbf{w}_1, \mathbf{w}_2, \cdots, \mathbf{w}_N \in \mathbb{R}^{d_w}$. Region embeddings, also known as phrase embeddings or N-grams embeddings, have shown to be effective in various deep learning models for NLP tasks~\cite{wang2018learning,johnson2015effective,johnson2015semi,johnson2017deep,kim2014convolutional} due to the use of word order information. We apply appropriate zero paddings to keep the number of embeddings.


To obtain primary capsules $\mathbf{pc}_1, \mathbf{pc}_2, \cdots, \mathbf{pc}_I$ from region embeddings $\mathbf{w}_1, \mathbf{w}_2, \cdots, \mathbf{w}_N$, we propose to use a trainable multi-head attention layer. Many different attention layers have been studied for NLP tasks~\cite{bahdanau2014neural,vaswani2017attention,yang2016hierarchical,wu2016google}. Yang et al.~\cite{yang2016hierarchical} used a learnable head vector, instead of a vector from another sources, to attend different positions. Multi-head self-attention was introduced in Vaswani et al.~\cite{vaswani2017attention}, enabling multiple joint views of inputs. Inspired by these studies, we propose the trainable multi-head attention layer, which employ multiple trainable head vectors to perform the attention independently. Each head vector will lead to a primary capsule, which is supposed to represent one specific semantic meaning after training. In iCapsNets, we have $I$ different head vectors, corresponding to $I$ primary capsules. In the following sections, we will demonstrate how the trainable multi-head attention layer is suitable to work with the capsule layer and helps interpreting the primary capsules.

\subsection{Trainable Multi-head Attention Layer}\label{sec:attention}

In iCapsNets, the trainable multi-head attention layer actually transforms tensors into capsules~\cite{sabour2017dynamic}. To achieve such transformation from scalar-output feature detectors to vector-output capsules, prior studies~\cite{sabour2017dynamic,xiao2018mcapsnet,yang2018investigating} simply group convolutional scalar outputs into vectors to form primary capsules, in order to replicate learned knowledge across space and keep positional information. However, this transformation results in a large number of capsules encoding duplicate information as well as inactive capsules during the computation. It hurts the efficiency of the capsule layer. Moreover, as the total number of primary capsules increases as the spatial sizes of inputs increase, the number of training parameters in the capsule layer becomes excessive when the spatial sizes are large, as explained in Section~\ref{sec:capsule}. It prohibits the use of CapsNets on large-scale datasets, like those built by Zhang et al.~\cite{zhang2015character} and the ImageNet dataset~\cite{russakovsky2015imagenet}. Applying our trainable multi-head attention layer effectively addresses this problem, since the number of primary capsules is fixed as a hyperparameter of the model.


As illustrated in Figure~\ref{fig:architecture}, the trainable multi-head attention layer takes region embeddings $\mathbf{w}_1, \mathbf{w}_2, \cdots, \mathbf{w}_N \in \mathbb{R}^{d_w}$ as inputs. To produce $I$ primary capsules $\mathbf{pc}_1, \mathbf{pc}_2, \cdots, \mathbf{pc}_I \in \mathbb{R}^{d_p}$, there are $I$ trainable head vectors $\mathbf{h}_1, \mathbf{h}_2, \cdots, \mathbf{h}_I \in \mathbb{R}^{d_q}$. We name these head vectors as primary capsule queries. For each $\mathbf{h}_i \in \mathbb{R}^{d_q}, m \in [1, I]$, the attention mechanism determines region embeddings that are important to a specific sentence-level semantic meaning and aggregates them accordingly to form a primary capsule $ \mathbf{pc}_i$. Specifically, the attention procedure is
\begin{align}
	\textbf{for}\ & n = 1,2,\cdots,N\ \textbf{do} \nonumber\\
	&\mathbf{v}^i_n = \mathbf{W}^v_i \mathbf{w}_n \\
	&\mathbf{k}^i_n = \mathbf{W}^k_i \mathbf{w}_n \\
	&\alpha^i_n = \frac{\exp({\mathbf{h}_i^T\mathbf{k}^i_n/\sqrt{d_q}})}{\sum_{n'}\exp({\mathbf{h}_i^T\mathbf{k}^i_{n'}/\sqrt{d_q}})} \label{eqn:attn_weights}
\end{align}
where $\mathbf{W}^v_i \in \mathbb{R}^{d_p \times d_w}$ and $\mathbf{W}^k_i \in \mathbb{R}^{d_q \times d_w}$, and the aggregation is achieved by a weighted summation:
\begin{equation}
	\mathbf{pc}_i = \sum_{n} \alpha^i_n \mathbf{v}^i_n.
\end{equation}
Note that for one primary capsule query $\mathbf{h}_i$, $\mathbf{W}^v_i$ and $\mathbf{W}^k_i$ are shared for every region embedding. Here, $\mathbf{W}^v_i$ and $\mathbf{W}^k_i$ represent linear transformations that map region embeddings to a different embedding space for attention. Trained jointly with the primary capsule query $\mathbf{h}_i$, they are supposed to provide appropriate views of region embeddings with the focus on a specific semantic meaning. Consequently, $\mathbf{k}^i_n \in \mathbb{R}^{d_q}, n = 1,2,\cdots,N$ serve as the attention keys while $\mathbf{v}^i_n \in \mathbb{R}^{d_p}, n = 1,2,\cdots,N$ serve as the attention values that are used to form the primary capsule $\mathbf{pc}_i \in \mathbb{R}^{d_p}$. The coefficients $\alpha^i_n$ indicate whether $\mathbf{w}_n$ is informative in generating $\mathbf{pc}_i$. In Section~\ref{sec:interpreatation}, we use $\alpha^i_n$ to perform local interpretation.


Note that Eq.~\eqref{eqn:attn_weights} is equivalent to a \textit{Softmax} operation. And we can easily infer that
\begin{equation}\label{eqn:attn_constraint}
	\sum_{n}\alpha^i_n = 1,
\end{equation}
which indicates that in the attention mechanism, inputs compete with each other for their contributions to outputs. Intuitively, it means that only important parts of inputs go through the attention layer. Information that is irrelevant to the semantic meanings represented by primary capsules is discarded and only useful information is retained. In Section~\ref{sec:capsule}, we can see that, in the dynamic routing algorithm of the capsule layer, outputs compete with each other for receiving inputs. A class capsule gets activated only when receiving agreements from multiple active primary capsules. Our iCapsNets use the capsule layer after the attention layer, since they have complementary functionalities, \emph{i.e.}, the attention layer filters information and the capsule layer makes full use of the filtered information.

To conclude, using the trainable multi-head attention layer to transforms tensors into capsules is not only efficient but also technically sound. Moreover, our trainable multi-head attention layer provides a simple way to interpret primary capsules, which leads to global interpretability of iCapsNets. We illustrate the interpretation method in Section~\ref{sec:interpreatation}.

\subsection{Capsule Layer}\label{sec:capsule}

\begin{algorithm}[t]
	\caption{Dynamic Routing Algorithm~\cite{sabour2017dynamic}}\label{alg:dynamic_routing}
	\begin{algorithmic}[1]
		\STATE \textbf{procedure} ROUTING($\hat{\mathbf{pc}}_{j|i}$, $r$, $l$)
		\STATE for all capsule $i$ in layer $l$ and capsule $j$ in layer $(l+1)$: $b_{ij} \leftarrow 0$
		\FOR{$r$ iterations}
		\STATE for all capsule $i$ in layer $l$: $\beta_{ij} \leftarrow \frac{\exp(b_{ij})}{\sum_{j}\exp(b_{ij})}$ \label{line:caps_weights}
		\STATE for all capsule $j$ in layer $(l+1)$: $\mathbf{s}_j \leftarrow \sum_{i} \beta_{ij} \hat{\mathbf{pc}}_{j|i}$
		\STATE for all capsule $j$ in layer $(l+1)$: $\mathbf{cc}_j \leftarrow Squash (\mathbf{s}_j)$
		\STATE for all capsule $i$ in layer $l$ and capsule $j$ in layer $(l+1)$: $b_{ij} \leftarrow b_{ij} + \hat{\mathbf{pc}}_{j|i} \cdot \mathbf{cc}_j$
		\ENDFOR
		\RETURN $\mathbf{cc}_j$
	\end{algorithmic}
\end{algorithm}

Algorithm~\ref{alg:dynamic_routing} shows the original dynamic routing algorithm~\cite{sabour2017dynamic}. The \textit{Squash} function is used to normalize the capsules:
\begin{equation}
	Squash(x)=\frac{||x||^2}{1+||x||^2}\frac{x}{||x||},
\end{equation}
where $x$ is a capsule, \emph{i.e.}, a vector.

Note that the inputs to the algorithm is not the original primary capsules. Before the routing, we perform linear transformations on primary capsules to produce ``prediction vectors''~\cite{sabour2017dynamic}. To be specific, for each pair of a primary capsule $\mathbf{pc}_i \in \mathbb{R}^{d_p}$ and a class capsule $\mathbf{cc}_j \in \mathbb{R}^{d_c}$, we compute
\begin{equation}\label{eqn:prediction_vectors}
	\hat{\mathbf{pc}}_{j|i} = \hat{\mathbf{W}}_{ij} \mathbf{pc}_i + \hat{\mathbf{b}}_{ij},
\end{equation}
where $\hat{\mathbf{W}}_{ij} \in \mathbb{R}^{d_c \times d_p}$ and $\hat{\mathbf{b}}_{ij} \in \mathbb{R}^{d_c}$. For $I$ primary capsules and $J$ class capsules, it results in $I \times J \times (d_c \times d_p + d_c)$ training parameters, which are excessive when the number of primary capsules is large. Using our trainable multi-head attention layer addresses this problem by limiting the number of primary capsules. However, a more direct solution is to have $\hat{\mathbf{W}}_{ij}$ shared across primary capsules, which means
\begin{equation}\label{eqn:weight_sharing}
	\hat{\mathbf{pc}}_{j|i} = \hat{\mathbf{W}}_{j} \mathbf{pc}_i + \hat{\mathbf{b}}_{ij},
\end{equation}
where $\hat{\mathbf{W}}_{j}$ is shared for every $\mathbf{pc}_i$. iCapsNets employ Eq.~\eqref{eqn:weight_sharing} to improve the efficiency. More importantly, we find that the weight sharing casts a regularization effect on primary capsule queries in our trainable multi-head attention layer, which is shown in Section~\ref{sec:visualization}.

\begin{figure}[t]
	\centering
	\includegraphics[width=0.45\textwidth]{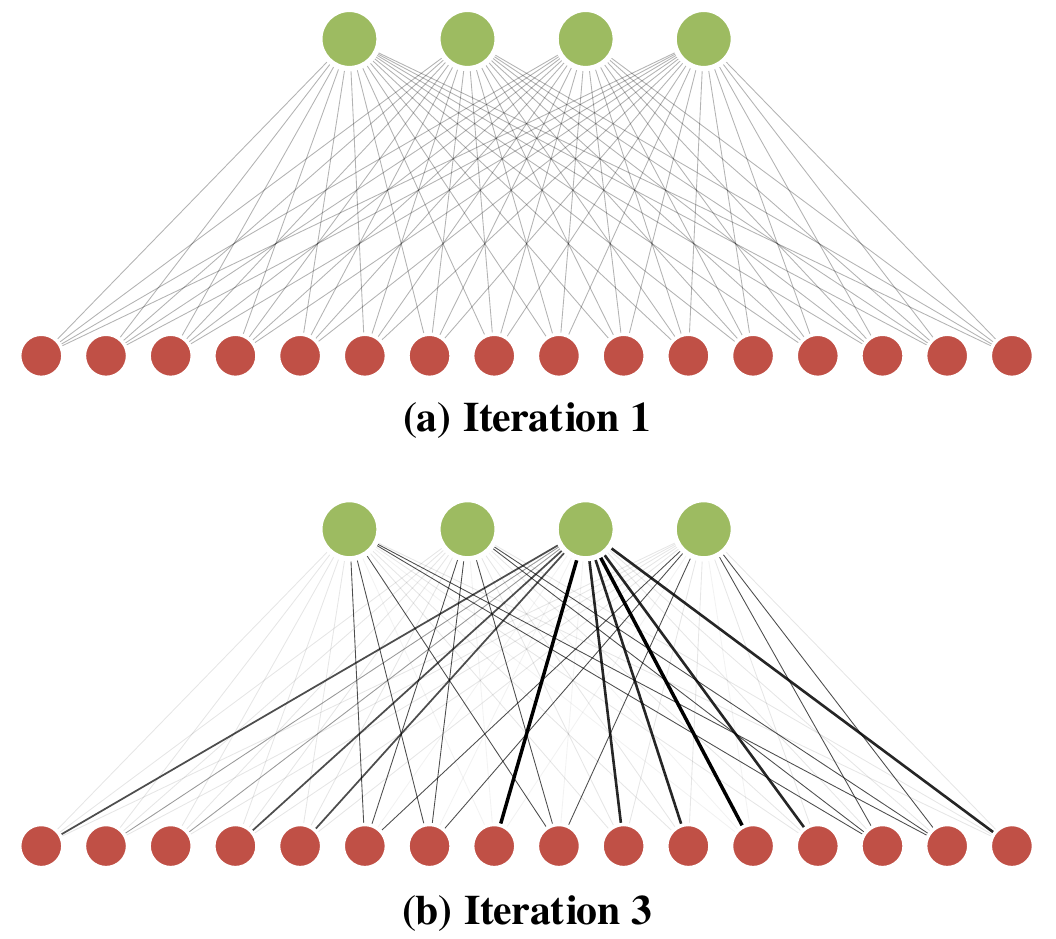}
	\caption{An illustration of the capsule layer. The visualization comes from an iCapsNet with 16 primary capsules trained on the AG's News dataset. Thicker lines indicate larger routing weights $\beta_{ij}$.}
	\label{fig:capsule}
\end{figure}

The dynamic routing algorithm computes weights $\beta_{ij}$ between every pair of a primary capsule $\mathbf{pc}_i$ and a class capsule $\mathbf{cc}_j$ in an iterative way. The process is visualized in Figure~\ref{fig:capsule}. By comparing the routing weights $\beta_{ij}$, we can see that, after $r=3$ iterations, a primary capsule may contribute much more to one of the class capsules than the others. This primary capsule is usually activated with a norm close to 1 and serves as a strong support to a class capsule. In another case, a primary capsule may contribute similarly to each class capsule. It means that either the primary capsule has a norm close to 0 or it captures a semantic meaning that is not helpful to classification. Therefore, it is reasonable to use the the routing weights $\beta_{ij}$ to explore the local interpretability of iCapsNets, as explained in Section~\ref{sec:interpreatation}.

To support the statement in Section~\ref{sec:attention}, we point out that line~\ref{line:caps_weights} in Algorithm~\ref{alg:dynamic_routing} corresponds to a \textit{Softmax} operation, which indicates
\begin{equation}\label{eqn:caps_constraint}
	\sum_{j}\beta_{ij} = 1,
\end{equation}
which is opposite to Eq.~\eqref{eqn:attn_constraint} in terms of normalizing weights across inputs or outputs. As mentioned above, Eq.~\eqref{eqn:attn_constraint} and Eq.~\eqref{eqn:caps_constraint} show a complementary relationship between the trainable multi-head attention layer and the capsule layer.

\subsection{iCapsNets for Long Documents}\label{sec:hierarchical_attention}

\begin{figure*}[t]
	\centering
	\includegraphics[width=\textwidth]{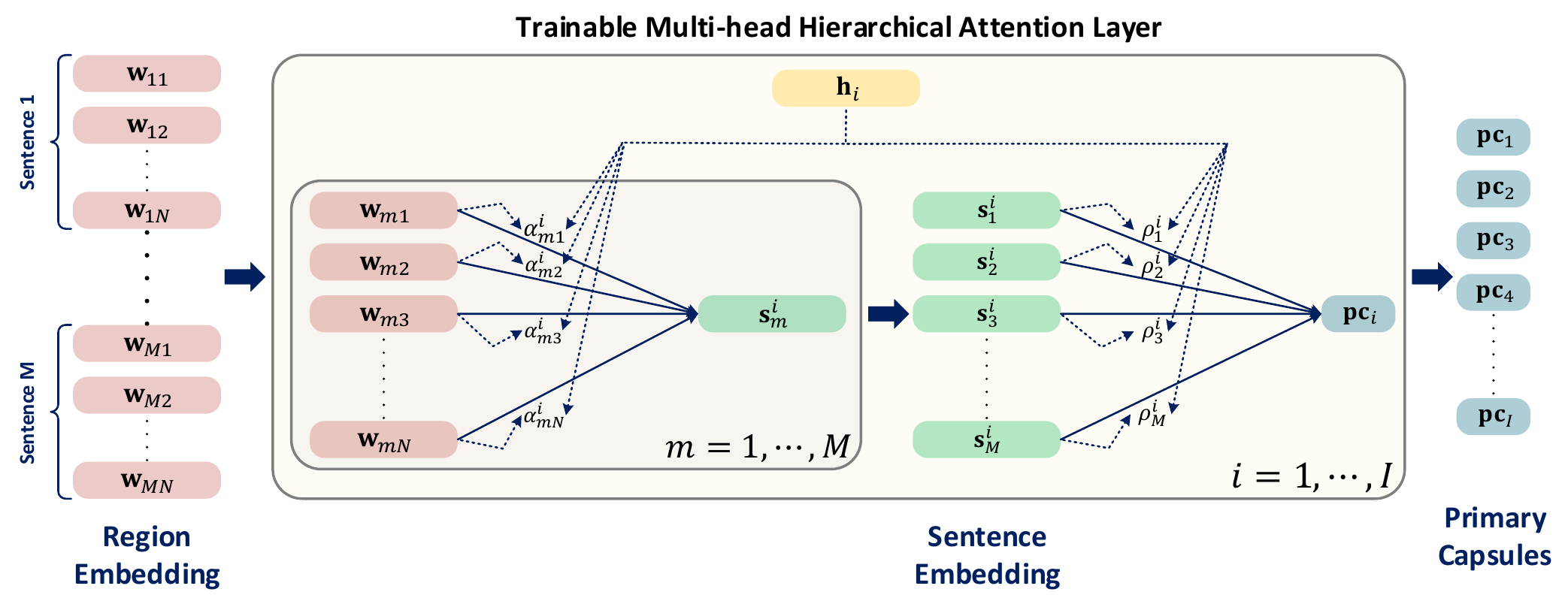}
	\caption{An illustration of the trainable multi-head hierarchical attention layer used in iCapsNets for long documents. Details are provided in Section~\ref{sec:hierarchical_attention}.}
	\label{fig:hierarchy}
\end{figure*}

In the discussions above, we assume that the input to iCapsNets is a single sentence of $N$ words, leading to a hierarchical word-region-sentence architecture. While it is reasonable for short documents, removing the assumption and adding a document level to the hierarchy usually results in a performance boost for long documents~\cite{tang2015document,yang2016hierarchical}. Thus, we propose the trainable multi-head hierarchical attention layer to adapt iCapsNets for long documents, as illustrated in Figure~\ref{fig:hierarchy}.

Consider a document of $M$ sentences, where each sentence has $N$ words. For each sentence, we use the same word embedding layer and 1-D convolutional layer to obtain region embeddings $\mathbf{w}_{mn} \in \mathbb{R}^{d_w}$, $m=1,2,\cdots,M$, $n=1,2,\cdots,N$. Our trainable multi-head hierarchical attention layer still has $I$ primary capsule queries $\mathbf{h}_1, \mathbf{h}_2, \cdots, \mathbf{h}_I \in \mathbb{R}^{d_q}$, corresponding to $I$ primary capsules $\mathbf{pc}_1, \mathbf{pc}_2, \cdots, \mathbf{pc}_I \in \mathbb{R}^{d_p}$. For each $\mathbf{h}_i$, it performs two levels of attention procedures:
\begin{align}
	\textbf{for}\ &m = 1,2,\cdots,M\ \textbf{do} \nonumber \\
	&\textbf{for}\ n = 1,2,\cdots,N\ \textbf{do} \nonumber \\
	&\ \ \ \ \ \ \ \mathbf{v}^i_{mn} = \mathbf{W}^{v}_i \mathbf{w}_{mn}, \ \mathbf{k}^i_{mn} = \mathbf{W}^{k}_i \mathbf{w}_{mn} \\
	&\ \ \ \ \ \ \ \alpha^i_{mn} = \frac{\exp({\mathbf{h}_i^T\mathbf{k}^i_{mn}/\sqrt{d_q}})}{\sum_{n'}\exp({\mathbf{h}_i^T\mathbf{k}^i_{mn'}/\sqrt{d_q}})} \\
	&\mathbf{s}^i_m = \sum_{n} \alpha^i_{mn} \mathbf{v}^i_{mn} \\
	&\tilde{\mathbf{v}}^i_m = \tilde{\mathbf{W}}^v_i \mathbf{s}^i_m, \ \tilde{\mathbf{k}}^i_m = \tilde{\mathbf{W}}^k_i \mathbf{s}^i_m \\
	&\rho^i_m = \frac{\exp({\mathbf{h}_i^T\tilde{\mathbf{k}}^i_m/\sqrt{d_q}})}{\sum_{m'}\exp({\mathbf{h}_i^T\tilde{\mathbf{k}}^i_{m'}/\sqrt{d_q}})} \\
	\mathbf{pc}_i &= \sum_{m} \rho^i_m \tilde{\mathbf{v}}^i_m
\end{align}
where $\mathbf{s}^i_{m} \in \mathbb{R}^{d_s}$, $\mathbf{W}^v_i \in \mathbb{R}^{d_s \times d_w}$, $\mathbf{W}^k_i \in \mathbb{R}^{d_q \times d_w}$, $\tilde{\mathbf{W}}^v_i \in \mathbb{R}^{d_p \times d_s}$, and $\tilde{\mathbf{W}}^k_i \in \mathbb{R}^{d_q \times d_s}$. Basically, we first apply the same attention layer on each sentence independently and obtain $M$ sentence embeddings $\mathbf{s}^i_{1}, \mathbf{s}^i_{2}, \cdots, \mathbf{s}^i_{M} \in \mathbb{R}^{d_s}$. These sentence embeddings focus on the semantic meaning that $\mathbf{pc}_i$ aims to capture. Next, we use an attention layer on $\mathbf{s}^i_{1}, \mathbf{s}^i_{2}, \cdots, \mathbf{s}^i_{M}$ to determine which sentences are more informative and aggregate them to produce $\mathbf{pc}_i$. The same procedure is applied for each $\mathbf{h}_i$. Note that in the two levels of attention procedures, we employ the same set of trainable head vectors $\mathbf{h}_1, \mathbf{h}_2, \cdots, \mathbf{h}_I$.

We denote the iCapsNets in Figure~\ref{fig:architecture} as iCapsNets$_{Short}$ and the ones with the trainable multi-head hierarchical attention layer as iCapsNets$_{Long}$. In the experiments, iCapsNets$_{Long}$ achieve significantly better accuracies for tasks on long documents.

\subsection{Interpreting iCapsNets}\label{sec:interpreatation}

\begin{figure}[t]
	\centering
	\includegraphics[width=0.45\textwidth]{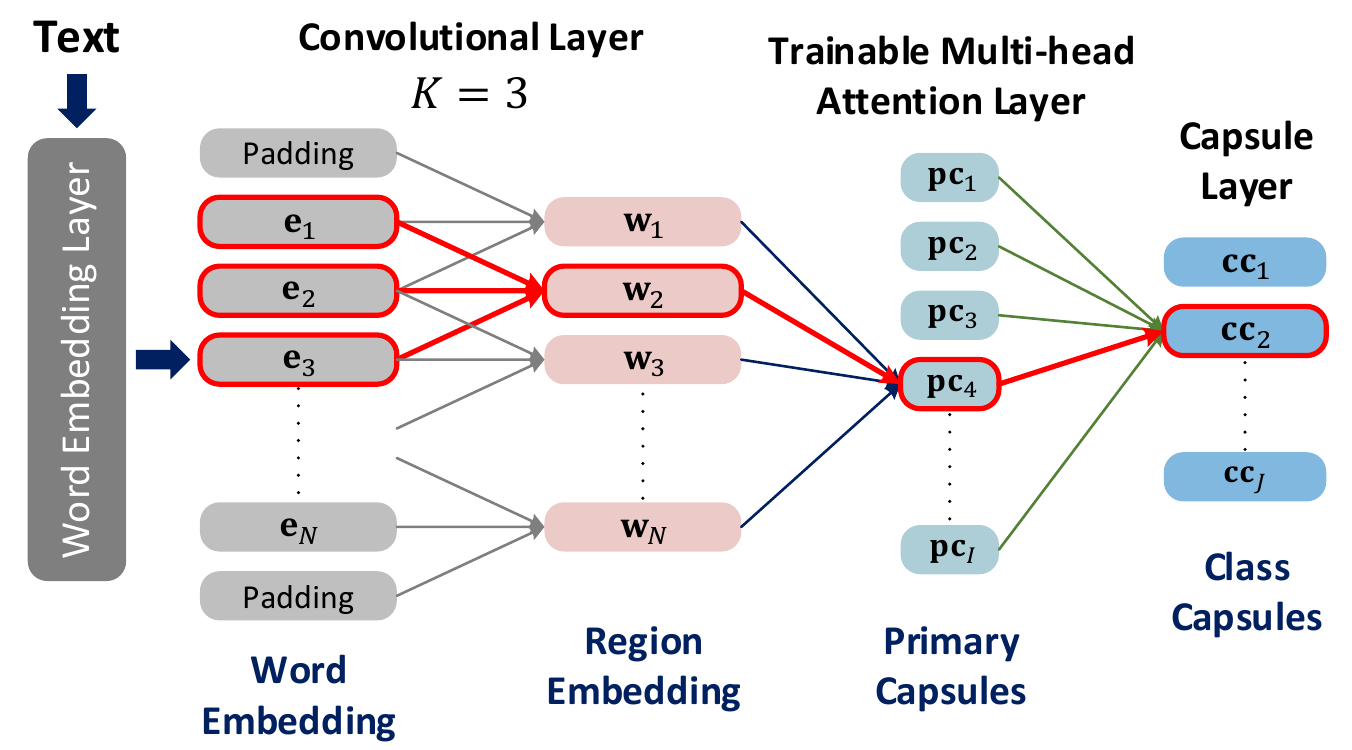}
	\caption{An illustration of the local interpretation process of iCapsNets. In this example, $k_1 = k_2 = 1$ and $K=3$. Details are provided in Section~\ref{sec:interpreatation}.}
	\label{fig:interpretation}
\end{figure}

iCapsNets can be interpreted both globally and locally~\cite{du2018techniques}. We first describe how to perform the local interpretation, \emph{i.e.}, explaining why the classification is made given a specific data sample. Note that, for the trainable multi-head attention layer and the capsule layer, each output is a weighted sum of all inputs, leading to a fully-connected pattern between inputs and outputs. However, unlike regular fully-connected layers where weights are fixed after training, weights in the trainable multi-head attention layer and the capsule layer are input-dependent. This property is crucial to achieving good local interpretation results. Specifically, after feeding a data sample into iCapsNets, we can extract parts of inputs that are important to the classification results, by simply examining large weights in iCapsNets. An illustration of the local interpretation process of iCapsNets is provided in Fig.~\ref{fig:interpretation}. Suppose iCapsNets categorize a data sample into class $j \in [1, J]$. We first obtain the top $k_1$ largest values from routing weights $\beta_{1j}, \beta_{2j}, \cdots, \beta_{Ij}$ in the capsule layer. As explained in Section~\ref{sec:capsule}, a large $\beta_{ij}$ means that the $i$-th primary capsule strongly support the $j$-th class capsule. Thus, the top $k_1$ largest weights indicate that the corresponding $k_1$ primary capsules are important to the prediction. Next, for each of these $k_1$ sentence capsules, we check the top $k_2$ largest values among weights in the trainable multi-head attention layer. For example, if $\mathbf{pc}_i$ is one of the $k_1$ primary capsules, we select the top $k_2$ largest values from $\alpha^i_1, \alpha^i_2, \cdots, \alpha^i_N$. As pointed out in Section~\ref{sec:attention}, large weights indicate that the corresponding inputs are informative in generating $\mathbf{pc}_i$. The process gives us $k_1 \times k_2$ $K$-grams, serving as the explanation for the classification result with respect to the input data sample. Here, $K$ is the kernel size of the 1-D convolutional layer. These $K$-grams may have overlapping words. The more number of times a word appears, the more important that word is. For iCapsNets$_{Long}$ with the trainable multi-head hierarchical attention layer, the local interpretation process is similar. For an important $\mathbf{pc}_i$, we first select the largest weight from $\rho^i_1, \rho^i_2, \cdots, \rho^i_M$, say $\rho^i_{m^*}$. Then we pick the top $k_2$ largest values from $\alpha^i_{m^*1}, \alpha^i_{m^*2}, \cdots, \alpha^i_{m^*N}$. The remaining parts are the same.

The global interpretation means interpreting semantically meaningful components in the model. It demonstrates how the model works generally with respect to the whole dataset. In iCapsNets, we attempt to determine the semantic meanings captured by each primary capsule to explore the global interpretability. First, we count the number of times when a primary capsule has the largest routing weight to the class capsule corresponding to the predicted class. Concretely, we maintain a frequency matrix $\mathbf{C}=[c_{ji}] \in \mathbb{N}^{J \times I}$ where $c_{ji}$ is initialized to be 0. For every data sample in the testing dataset, we performs the local interpretation described above with $k_1=k_2=1$. If it is classified into the $j$-th class and $\beta_{ij}$ is the largest routing weight among $\beta_{1j}, \beta_{2j}, \cdots, \beta_{Ij}$, we let $c_{ji} \leftarrow c_{ji} + 1$. Meanwhile, for each $c_{ij}$, we maintain a list of words in the resulted $K$-gram. We find the simple statistical method gives good global interpretation results, as shown in Section~\ref{sec:global_interpret_results}. The final $\mathbf{C}$ shows a sparse pattern; that is, only a few values in $\mathbf{C}$ are large. And the most frequent words can indicate the semantic meaning captured by primary capsules.

Both local and global interpretabilities of iCapsNets are achieved using simple methods. In the experiments, iCapsNets show a good trade-off between accuracy and interpretability.

\begin{table*}[t]
	\caption{Statistics of the 7 large-scale datasets from Zhang et al.~\cite{zhang2015character}.}
	\begin{center}
		\begin{tabular}{l|crrcc}
			Dataset & Classes & Train Samples & Test Samples & Avg. Lengths & Tasks \\
			\hline
			Yelp Review Polarity & 2 & 560,000 & 38,000 & 156 & Sentiment \\
			Yelp Review Full & 5 & 650,000 & 50,000 & 158 & Sentiment \\
			Yahoo! Answers & 10 & 1,400,000 & 60,000 & 112 & Topic \\
			Amazon Review Polarity & 2 & 3,600,000 & 400,000 & 91 & Sentiment \\
			Amazon Review Full & 5 & 3,000,000 & 650,000 & 93 & Sentiment \\
			AG's News & 4 & 120,000 & 7,600 & 44 & Topic \\
			DBPedia & 14 & 560,000 & 70,000 & 55 & Ontology \\
		\end{tabular}
		\label{table:dataset}
	\end{center}
\end{table*}

\section{Experimental Studies}

We perform thorough experiments to evaluate and analyze iCapsNets. First, we demonstrate the local and global interpretability of iCapsNets. Then, in terms of classification accuracy, we compare iCapsNets with several text classification baselines which are not interpretable. Notably, iCapsNets achieve a good trade-off between interpretability and accuracy.

\subsection{Datasets}

We conduct experiments on 7 publicly available large-scale datasets built by Zhang et al.~\cite{zhang2015character}. These datasets cover different text classification tasks, such as sentiment analysis, topic categorization, and ontology extraction. Table~\ref{table:dataset} is a detailed summary of these datasets. In particular, without loss of generality, we demonstrate the local and global interpretability of iCapsNets using examples from the AG's News dataset and the Yahoo! Answers dataset.

\begin{table*}[!t]
	\caption{Hyperparameter settings of iCapsNets$_{Short}$ on the 7 datasets from Zhang et al.~\cite{zhang2015character}. Explanations of these hyperparameters are provided in Section~\ref{appen:1}.}
	\begin{center}
		\begin{tabular}{l|cccccccc}
			Dataset & $V$ & $F$ & $d_e$ & $K$ & $d_w$ & $d_q=d_p$ & $d_c$ & $N$ \\
			\hline
			Yelp Review Polarity & 25,102 & 50 & 300+64 & 5 & 512 & 16 & 32 & 1,296 \\
			Yelp Review Full & 27,729 & 50 & 300+64 & 5 & 512 & 16 & 32 & 1,438 \\
			Yahoo! Answers & 35,194 & 100 & 300+64 & 5 & 512 & 16 & 32 & 1,000 \\
			Amazon Review Polarity & 33,207 & 200 & 300+64 & 5 & 512 & 16 & 32 & 592 \\
			Amazon Review Full & 17,534 & 500 & 300+64 & 5 & 512 & 16 & 32 & 592 \\
			AG's News & 30,794 & 5 & 300+32 & 3 & 256 & 8 & 16 & 195 \\
			DBPedia & 26,141 & 50 & 300+64 & 5 & 256 & 8 & 16 & 1,588
		\end{tabular}
		\label{table:short_setup}
	\end{center}
\end{table*}

\begin{table*}[!t]
	\caption{Hyperparameter settings of iCapsNets$_{Long}$ on the 7 datasets from Zhang et al.~\cite{zhang2015character}. Explanations of these hyperparameters are provided in Section~\ref{appen:1}.}
	\begin{center}
		\begin{tabular}{l|ccccccccc}
			Dataset & $V$ & $F$ & $d_e$ & $K$ & $d_w$ & $d_q=d_s=d_p$ & $d_c$ & $M$ & $N$ \\
			\hline
			Yelp Review Polarity & 77,202 & 5 & 300+64 & 5 & 512 & 16 & 32 & 20 & 100 \\
			Yelp Review Full & 82,814 & 5 & 300+64 & 5 & 512 & 16 & 32 & 20 & 100 \\
			Yahoo! Answers & 131,081 & 10 & 300+32 & 5 & 512 & 16 & 32 & 15 & 100 \\
			Amazon Review Polarity & 155,192 & 10 & 300+64 & 5 & 512 & 16 & 32 & 15 & 100 \\
			Amazon Review Full & 142,375 & 10 & 300+64 & 5 & 512 & 16 & 32 & 15 & 100 \\
			AG's News & 30,794 & 5 & 300+32 & 3 & 256 & 8 & 16 & 10 & 86 \\
			DBPedia & 26,141 & 50 & 300+64 & 5 & 256 & 8 & 16 & 10 & 100 \\
		\end{tabular}
		\label{table:long_setup}
	\end{center}
\end{table*}

\section{Experimental Setups}\label{appen:1}

We introduce detailed experimental setups for reproducibility. Code is also provided~\footnote{https://www.dropbox.com/s/ev06l6x7ddy9pgb/iCapsNet.zip}.

The word embedding layer of iCapsNets involves the step of generating a vocabulary. The size of the vocabulary $V$ is determined by the training set and a minimum frequency $F$. Specifically, if a word appears more than $F$ times in the training set, it is included in the vocabulary. In iCapsNets, each word embedding is composed of two parts. The first part is the 300-dimensional pre-trained word2vec~\cite{mikolov2013distributed} and is fixed during training. The second part has dimension $(d_e-300)$ and is randomly initialized and trained. The 1-D convolutional layer with a kernel size of $K$ transforms the $d_e$-dimensional word embeddings into $d_w$-dimensional region embeddings. For the trainable multi-head attention layer in iCapsNets$_{Short}$, we let the dimension of each primary capsule query be equal to that of each primary capsule, \emph{i.e.}, $d_q = d_p$. For the trainable multi-head hierarchical attention layer in iCapsNets$_{Long}$, the dimension of the intermediate sentence embeddings is also set to be equal to $d_p$, \emph{i.e.}, $d_q = d_s = d_p$. In addition, we set the number of primary capsules $I$ to be $d_w / d_p$. The number of class capsules $J$ depends on the number of classes. The dimension of a class capsule is $d_c$. For iCapsNets$_{Short}$, the input is a single sentence. We use zero paddings to make all the inputs have the same number of words $N$ for large batch training. For iCapsNets$_{Long}$, the input is a document. We also apply zero paddings so that all the inputs have the same number of sentences $M$ and each sentence has the same number of words $N$. Table~\ref{table:short_setup} and~\ref{table:long_setup} provide our best settings of these hyperparameters of iCapsNets$_{Short}$ and iCapsNets$_{Long}$ for each dataset, respectively.

As the outputs of iCapsNets are class capsules, the predictions are made based on their norms. That is, the class capsule with the largest norm corresponds to the predicted class. To train iCapsNets, we apply the margin loss proposed by Sabour et al.~\cite{sabour2017dynamic}. To be specific, for each class capsule $\mathbf{cc}_j$, $j=1,2,\cdots,J$, a separate loss function is given by
\begin{equation}
\begin{aligned}
L_j = & I_{j^*}(j)\max(0, m^+-||\mathbf{cc}_j||)^2 \\
 &+ \lambda(1-I_{j^*}(j))\max(0, ||\mathbf{cc}_j||-m^-)^2,
\end{aligned}
\end{equation}
where $m^+=0.9$, $m^-=0.1$, $\lambda=0.5$, and $I_{j^*}(j)$ is an indicator function defined as
\begin{equation}
I_{j^*}(j)=
\begin{cases}
1,\quad & \textbf{if}\ j = j^* \\
0,\quad & \textbf{if}\ j \neq j^*
\end{cases},
\end{equation}
where $j^*$ is the index of the true label. The total loss is the sum of the loss function of all the class capsules. With the margin loss, the Adam optimizer is used to train iCapsNets. For iCapsNets$_{Short}$, the learning rate is set to $0.0001$ for the AG's News and DBPedia datasets and $0.001$ for the other 5 datasets. For iCapsNets$_{Long}$, the learning rate is set to $0.0005$ for all the 7 dataset.

\begin{figure}[t]
	\centering
	\frame{\includegraphics[width=0.45\textwidth]{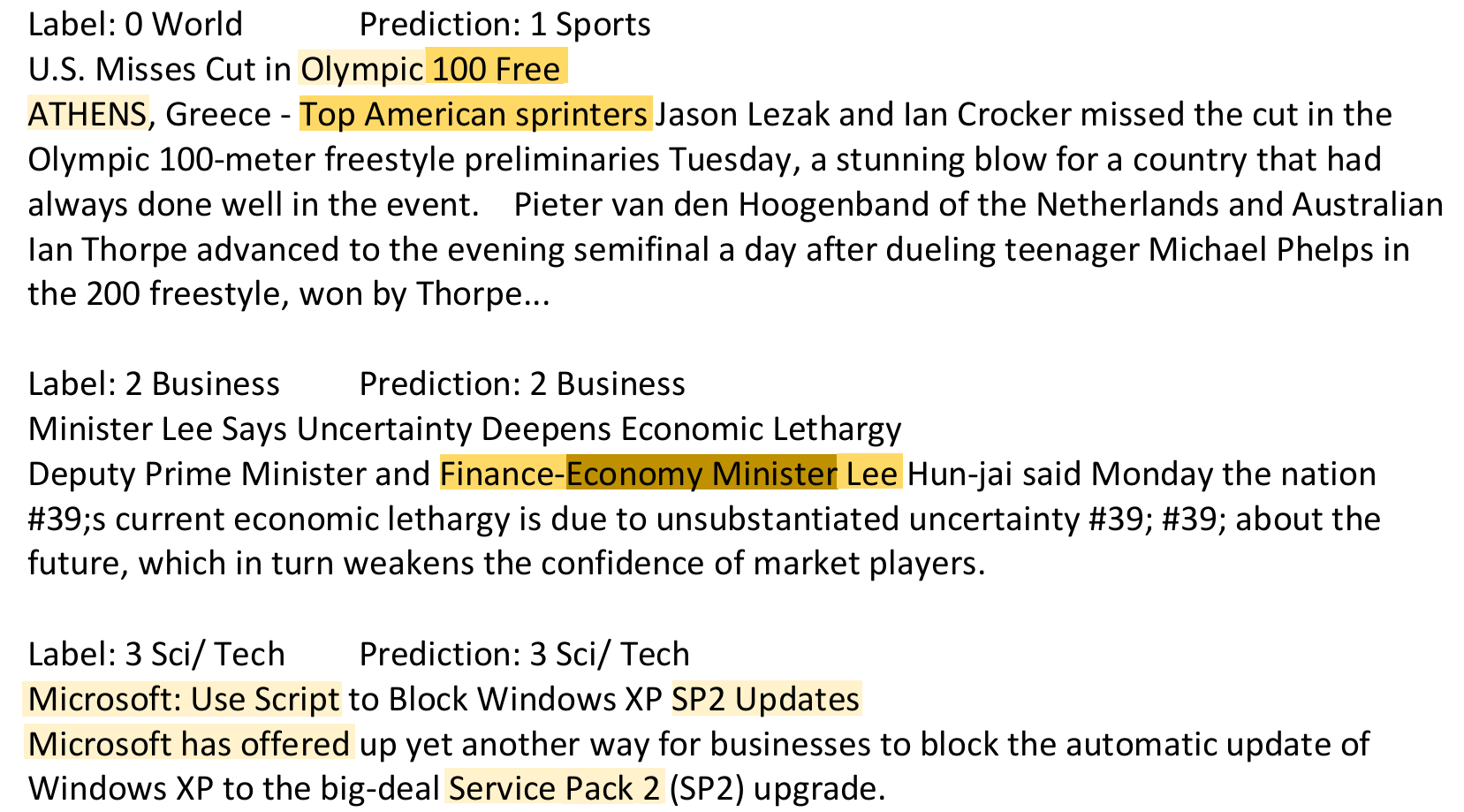}}
	\caption{Examples of local interpretation results of iCapsNets$_{Short}$ on the AG's News dataset. We set $k_1=k_2=2$ and $K=3$. Darker colors indicate more overlapping.}
	\label{fig:interpretag}
\end{figure}

\begin{figure}[t]
	\centering
	\frame{\includegraphics[width=0.45\textwidth]{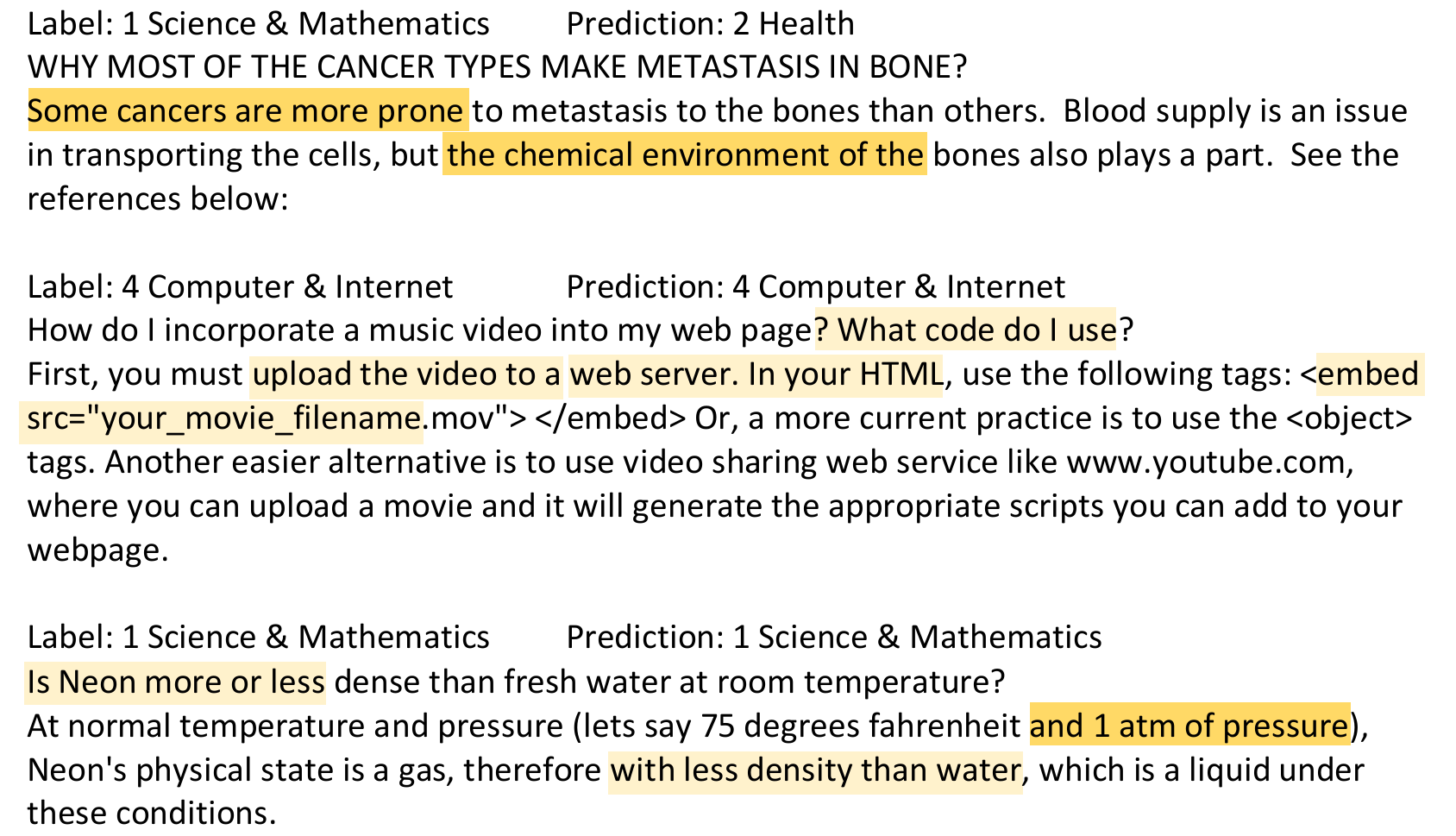}}
	\caption{Examples of local interpretation results of iCapsNets$_{Long}$ on the Yahoo! Answers dataset. We set $k_1=k_2=2$ and $K=5$. Darker colors indicate more overlapping.}
	\label{fig:interpretyahoo}
\end{figure}

\subsection{Local Interpretation Results}\label{sec:local_interpret_results}

\begin{figure*}[t]
	\centering
	\includegraphics[width=\textwidth]{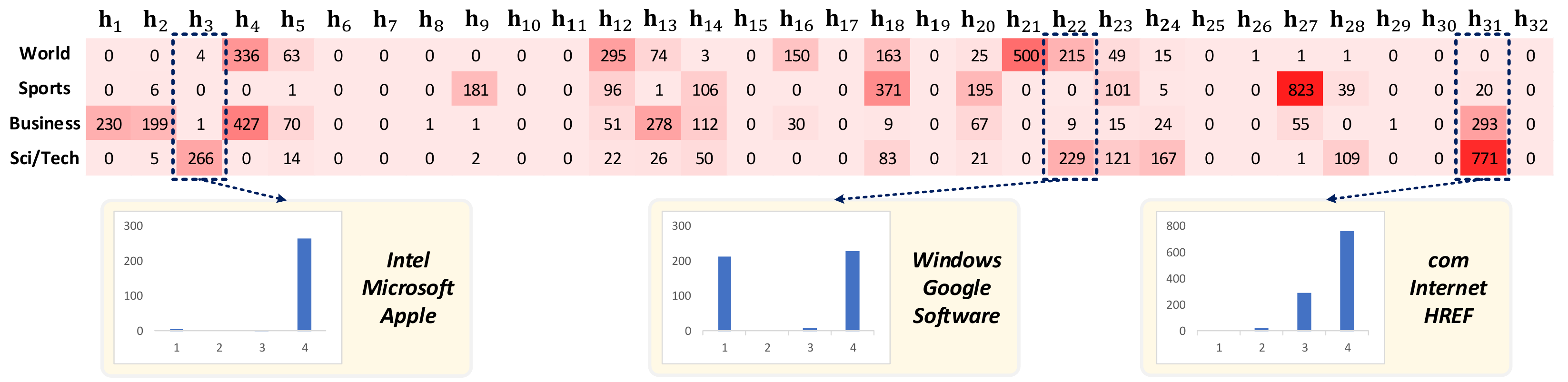}
	\caption{Visualization of the frequency matrix $\mathbf{C}$ for the global interpretation of iCapsNets$_{Short}$ on the AG's News dataset. For selected columns, we make a histogram and list the most frequent words as the interpretation of corresponding primary capsules. Note that primary capsules and primary capsule queries have a one-to-one relationship.}
	\label{fig:ag_heatmap}
\end{figure*}

\begin{figure*}[t]
	\centering
	\includegraphics[width=\textwidth]{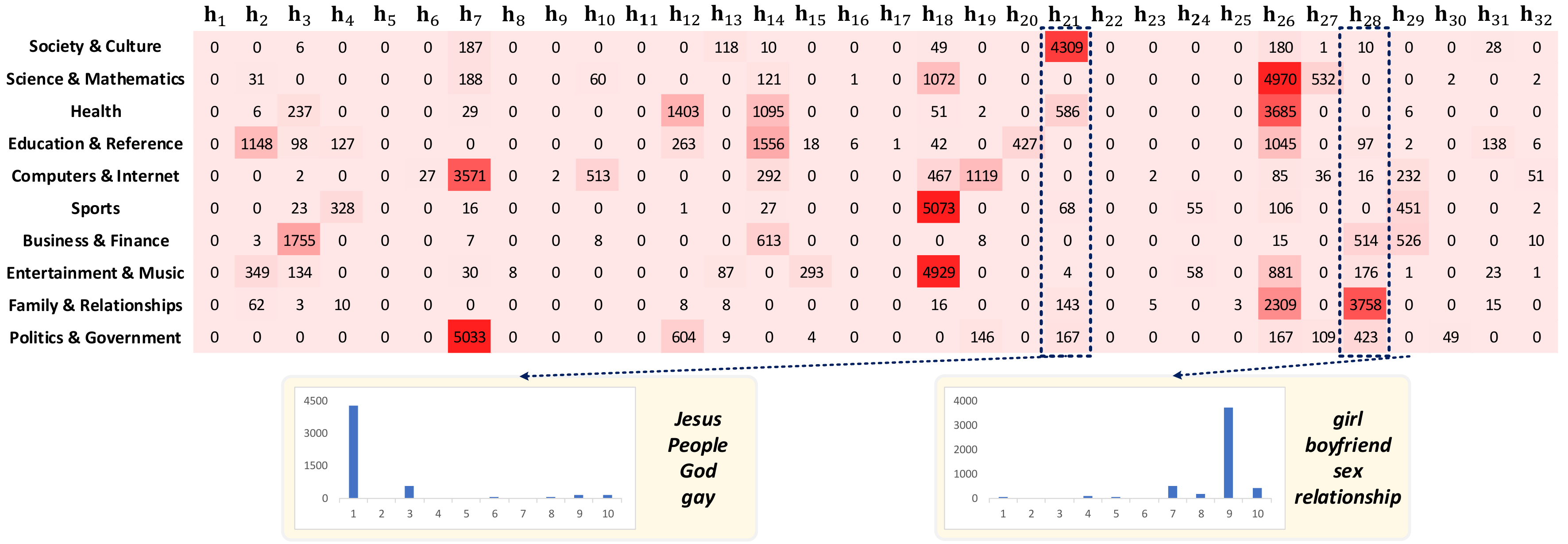}
	\caption{Visualization of the frequency matrix $\mathbf{C}$ for the global interpretation of iCapsNets$_{Long}$ on the Yahoo! Answers dataset. For selected columns, we make a histogram and list the most frequent words as the interpretation of corresponding primary capsules. Note that primary capsules and primary capsule queries have a one-to-one relationship.}
	\label{fig:yahoo_heatmap}
\end{figure*}

In order to demonstrate the local interpretability of iCapsNets, we show concrete examples of local interpretation results obtained through the interpretation method as described in Section~\ref{sec:interpreatation}. Specifically, we train iCapsNets$_{Short}$ on the AG's News dataset and iCapsNets$_{Long}$ on the Yahoo! Answers dataset, respectively. Then we take examples from the testing set to perform prediction and local prediction.

Figures~\ref{fig:interpretag} and~\ref{fig:interpretyahoo} provide examples of local interpretation results of iCapsNets on AG's News and Yahoo! Answers datasets, respectively. We can observe that, for each data sample, the extracted words well justify why iCapsNets make the predictions, no matter the predictions are correct or not.

\subsection{Global Interpretation Results}\label{sec:global_interpret_results}

We perform global interpretation for the same iCapsNets$_{Short}$ on the AG's News dataset and iCapsNets$_{Long}$ on the Yahoo! Answers dataset, respectively. To be specific, we first visualize the frequency matrix $\mathbf{C}$ as introduced in Section~\ref{sec:interpreatation}. The visualizations are provided in Figures~\ref{fig:ag_heatmap} and~\ref{fig:yahoo_heatmap}. We can see that in both cases, $\mathbf{C}$ shows a sparse pattern. Note that the $i$-th column corresponds to the primary capsule $\mathbf{pc}_i$, or equivalently the primary capsule query $\mathbf{h}_i$. Thus, for $\mathbf{pc}_i$, we can check the lists of frequent words corresponding to $c_{i1}, c_{i2}, \cdots, c_{iJ}$ and use the most frequent words to interpret $\mathbf{pc}_i$. Interpreting primary capsules leads to explanation on how iCapsNets work generally, \textit{i.e.} achieving the global interpretability.

\subsubsection{Visualization of Primary Capsule Queries}\label{sec:visualization}

\begin{table*}[t]
	\caption{Comparisons between different models in terms of test accuracies [\%] on the 7 datasets from Zhang et al.~\cite{zhang2015character}. iCapsNets achieve competitive results compared to non-interpretable models.}
	\begin{center}
		\tabcolsep=0.5cm\begin{tabular}{l|ccccccc}
			Model & Yelp P. & Yelp F. & Yah. A. & Amz P. & Amz F. & AG & DBP \\
			\hline
			word-CNN~\cite{zhang2015character} & 95.4 & 60.4 & 71.2 & 94.5 & 57.6 & 91.5 & 98.6 \\
			char-CNN~\cite{zhang2015character} & 95.1 & 62.1 & 71.2 & 94.5 & 59.6 & 90.5 & 98.5 \\
			char-CNN+RNN~\cite{xiao2016efficient} & 94.5 & 61.8 & 71.7 & 94.1 & 59.2 & 91.4 & 98.6 \\
			char-VDCNN~\cite{conneau2017very} & 95.7 & 64.7 & 73.4 & \textbf{95.7} & \textbf{63.0} & 91.3 & 98.7 \\
			D-LSTM~\cite{yogatama2017generative} & 92.6 & 59.6 & 73.7 & - & - & 92.1 & 98.7 \\
			\hline
			FastText~\cite{joulin2017bag} & 95.7 & 63.9 & 72.3 & 94.6 & 60.2 & 92.5 & 98.6 \\
			W.C.region.emb~\cite{qiao2018a} & \textbf{96.4} & \textbf{64.9} & 73.7 & 95.1 & 60.9 & 92.8 & 98.9 \\
			C.W.region.emb~\cite{qiao2018a} & 96.2 & 64.5 & 73.4 & 95.3 & 60.8 & 92.8 & 98.9 \\
			\hline
			CapsNets~\cite{yang2018investigating} & - & - & - & - & - & 92.6 & - \\
			\hline
			iCapsNets$_{Short}$ & 95.9 & 64.2 & 73.5 & 95.2 & 61.1 & \textbf{93.0} & \textbf{99.0} \\
			iCapsNets$_{Long}$ & 96.0 & 64.3 & \textbf{74.5} & \textbf{95.7} & 62.4 & 92.9 & 98.9 \\
		\end{tabular}
		\label{table:result}
	\end{center}
\end{table*}

\begin{figure}[t]
	\centering
	\includegraphics[width=0.45\textwidth]{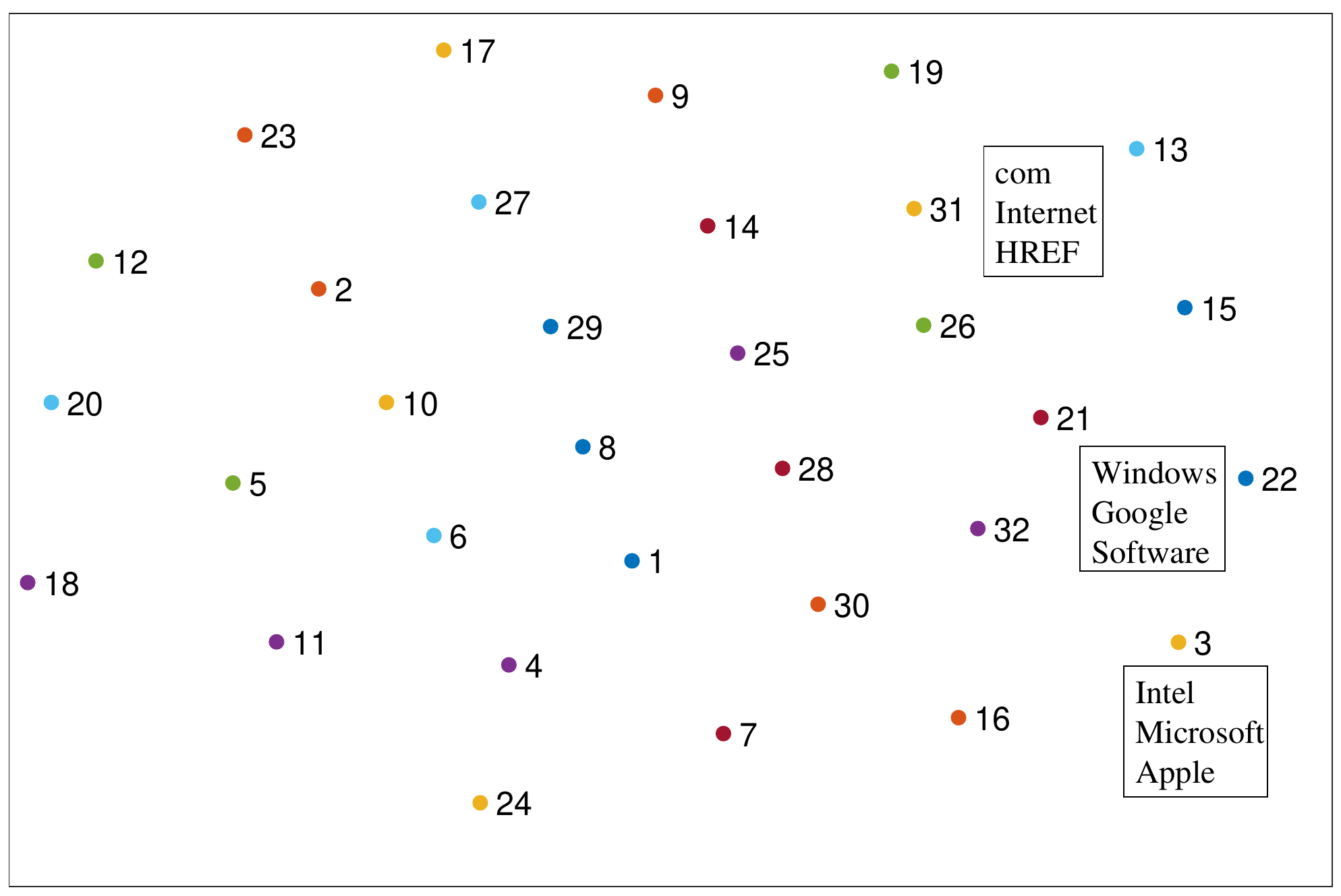}
	\caption{t-SNE visualization for the 32 primary capsule queries of iCapsNets$_{Short}$ on the AG's News dataset after training. The words come from Figure~\ref{fig:ag_heatmap}.}
	\label{fig:ag_vis}
\end{figure}

We further visualize the sparse pattern of the primary capsule queries in the embedding space. Specifically, we perform t-SNE visualization~\cite{maaten2008visualizing} of the primary capsule queries $\mathbf{h}_1, \mathbf{h}_2, \cdots, \mathbf{h}_{32}$. Fig.~\ref{fig:ag_vis} shows the visualization of iCapsNets$_{Short}$ trained on the AG's News dataset. It is observed that the primary capsule queries distribute sparsely on the plane, indicating that they capture different semantic meanings under the same embedding space.

\subsection{Classification Results}

Last, we show that iCapsNets are able to achieve competitive results compared to non-interpretable models.

\subsubsection{Baselines}

We select several popular supervised text classification models as baselines to show that iCapsNets can achieve competitive accuracies. In terms of deep learning models, we compare iCapsNets with the word-level convolutional model (word-CNN)~\cite{kim2014convolutional} and character-level convolutional model (char-CNN)~\cite{zhang2015character}. Comparisons with two variants of char-CNN, the character-level convolutional recurrent model (char-CNN+RNN)~\cite{xiao2016efficient} and the very deep character-level convolutional model (char-VDCNN)~\cite{conneau2017very}, are also conducted. In addition, iCapsNets are compared with the discriminative LSTM model (D-LSTM)~\cite{yogatama2017generative}. FastText~\cite{joulin2017bag} combines distributed representations of words with traditional model BoW and gets improved by using the word-context region embeddings (W.C.region.emb) and context-word region embeddings (C.W.region.emb) \cite{qiao2018a}. We report the accuracies of these baselines from Zhang et al.~\cite{zhang2015character} and Qiao et al.~\cite{qiao2018a}. As iCapsNets are based on CapsNets~\cite{sabour2017dynamic}, we also include CapsNets as baselines. CapsNets have been investigated on text classification~\cite{yang2018investigating}. However, due to the efficiency problem discussed in Section~\ref{sec:attention}, original CapsNets can only work well on small datasets. Therefore, only the accuracy on the AG's News dataset is available.

\subsubsection{Results}

Table~\ref{table:result} summarizes the classification results of all models. In terms of test accuracies, iCapsNets outperforms all the baselines on 4 of the 7 datasets. On the other 3 datasets, iCapsNets achieve competitive results.

Deep learning models based on RNNs, like char-CNN+RNN and D-LSTM, are usually hard to interpret as RNNs process texts sequentially and do not tell which parts of the sequence are informative. The interpretability of CNN models with word embeddings has been studied~\cite{yuan2019interpreting}. However, the interpretation process is computational expensive. And only local interpretability has been explored. Applying character embeddings usually improve the accuracies. However, as pointed out by Conneau et al.~\cite{conneau2017very}, the models may process a sentence as a stream of signals, which we can not understand semantically.

FastText, W.C.region.emb, and C.W.region.emb combine distributed representations with traditional model BoW. They are efficient and effective on text classification tasks. However, an average/sum operation is employed to generate sentence embeddings from word or region embeddings, making the model not interpretable.

As interpretable models usually suffer from the performance loss~\cite{du2018techniques}, the classification performance of iCapsNets is strong considering its interpretability.

\section{Conclusions}

In this work, we aim to develop a deep learning model that achieves a good trade-off between accuracy and interpretability on text classification tasks. Based on our insights on capsules, we propose the interpretable capsule network~(iCapsNets) by employing attention mechanism and adapting CapsNets~\cite{sabour2017dynamic} from computer vision tasks to text classification tasks. We provide novel, simple yet effective way to interpret our iCapsNets. In particular, iCapsNets achieve the local and global interpretability at the same time. Experimental results show that our iCapsNets yield human-understandable interpretation results, without suffering from significant performance loss compared to non-interpretable models.

\ifCLASSOPTIONcompsoc
  \section*{Acknowledgments}
\else
  \section*{Acknowledgment}
\fi

This work was supported in part by National Science Foundation grant IIS-1908198 and Defense Advanced Research Projects Agency grant N66001-17-2-4031.

\ifCLASSOPTIONcaptionsoff
  \newpage
\fi



\bibliographystyle{IEEEtran}
\bibliography{reference}
\end{document}